\documentclass{article}

\usepackage[final,nonatbib]{neurips_2019}

\usepackage[utf8]{inputenc} % allow utf-8 input
\usepackage[T1]{fontenc}    % use 8-bit T1 fonts
\usepackage{hyperref}       % hyperlinks
\usepackage{url}            % simple URL typesetting
\usepackage{booktabs}       % professional-quality tables
\usepackage{amsfonts}       % blackboard math symbols
\usepackage{nicefrac}       % compact symbols for 1/2, etc.
\usepackage{microtype}      % microtypography

\usepackage{booktabs}       % professional-quality tables
\usepackage{amsfonts}       % blackboard math symbols
\usepackage{microtype}      % microtypography
\usepackage{xcolor}
\usepackage{url}
\usepackage{verbatim} % allows multiline comments
\usepackage{graphicx}
\usepackage{multirow}
\usepackage{algorithm}
\usepackage[noend]{algorithmic}
\usepackage{xspace}
\usepackage{epsfig}
\usepackage{amsmath}
\usepackage{amsthm}
\usepackage{amssymb}
\usepackage{times}
\usepackage{xr}
\usepackage{bbm}
\usepackage{bm}
\usepackage{subcaption}
\usepackage{enumitem}
\usepackage{hyperref}       % hyperlinks
\usepackage{multicol}
\usepackage{tabularx}
\usepackage{perpage} %the perpage package
\MakePerPage{footnote} %the perpage package command
\usepackage{wrapfig}

\newcommand{\xhdr}[1]{{\noindent\bfseries #1}.}

\newcommand{\mb}{\mathbf}
\newcommand{\cut}[1]{}

\newtheorem{definition}{Definition}

\newtheorem{observation}{\textbf{Observation}}

\DeclareMathOperator*{\argmax}{argmax}
\DeclareMathOperator{\Bernoulli}{Bernoulli}
\DeclareMathOperator{\Multimonial}{Multimonial}
\DeclareMathOperator{\NodeMerge}{NodeMerge}
\DeclareMathOperator{\NodeSplit}{NodeSplit}
\DeclareMathOperator{\Uniform}{Uniform}
\DeclareMathOperator{\Degree}{Degree}
\DeclareMathOperator{\GCN}{GCN}

\newcommand{\name}{G2SAT}
\newcommand*{\affmark}[1][*]{\textsuperscript{#1}}
\newcommand*\samethanks[1][\value{footnote}]{\footnotemark[#1]}

\newcommand{\hide}[1]{}

\title{\name: Learning to Generate SAT Formulas}

\author{
Jiaxuan You\affmark[1]\thanks{The two first authors made equal contributions.} \\
\texttt{jiaxuan@cs.stanford.edu}
\And 
Haoze Wu\affmark[1]\samethanks\\
\texttt{haozewu@stanford.edu}
\And 
Clark Barrett\affmark[1] \\
\texttt{barrett@cs.stanford.edu}
\And 
Raghuram Ramanujan\affmark[2] \\
\texttt{raramanujan@davidson.edu}
\And 
Jure Leskovec\affmark[1] \\
\texttt{jure@cs.stanford.edu}
}

\begin{document}

\maketitle
\vspace{-2.5em}
 \begin{center}
 \affmark[1]Department of Computer Science, Stanford University \\ \affmark[2]Department of Mathematics and Computer Science, Davidson College\\
 \end{center}
 \vspace{1em}

\begin{abstract}
    The Boolean Satisfiability (SAT) problem is the canonical NP-complete problem and is fundamental to computer science, with a wide array of applications in planning, verification, and theorem proving. Developing and evaluating practical SAT solvers relies on extensive empirical testing on a set of real-world benchmark formulas. However, the availability of such real-world SAT formulas is limited. While these benchmark formulas can be augmented with synthetically generated ones, existing approaches for doing so are heavily hand-crafted and fail to simultaneously capture a wide range of characteristics exhibited by real-world SAT instances. %\raghu{The next three sentences appear a little redundant, consolidate?} 
    In this work, we present G2SAT, the first deep generative framework that learns to generate SAT formulas from a given set of input formulas. Our key insight is that SAT formulas can be transformed into latent bipartite graph representations which we model using a specialized deep generative neural network. %The core of G2SAT is a novel deep generative model for bipartite graphs, which generates graphs via iterative node merging operations.
    % SAT-GEN includes three phases. First, we represent an input SAT formulas as the corresponding Literal-Incidence Graphs (LIGs). Then, we train a specialized deep graph generative model to capture the complex joint distribution of all nodes and edges in LIGs, and the model can then be used to generate realistic LIGs. In the last phase, we develop a novel technique to project LIGs back to the SAT formulas, by finding a minimal clique edge cover of the given graphs using an efficient greedy hill-climbing algorithm. 
    %\jure{Update:}
    % We proposed two variants of \name, including LI\name~based on Literal-Incidence Graphs (LIGs) and LC\name~based on Literal-Clause Graphs (VCGs).
    We show that G2SAT can generate SAT formulas that closely resemble given real-world SAT instances, as measured by both graph metrics and SAT solver behavior. Further, we show that our synthetic SAT formulas could be used to improve SAT solver performance on real-world benchmarks, which opens up new opportunities for the continued development of SAT solvers and a deeper understanding of their performance.
    %Furthermore, we show that tuning the hyper-parameters of SAT solvers using the generated formulas can largely improve the solvers’ performance on real-world benchmarks, which opens up the opportunities for continued development of complete SAT-solvers and a deeper understanding of their performance.
\end{abstract}
\section{Introduction}
\label{sec:intro}

% \jiaxuan{what problem}
% \jiaxuan{why important}
The \emph{Boolean Satisfiability (SAT) problem} is central to computer science, and finds many applications across Artificial Intelligence, including planning \cite{planningAsSat}, verification \cite{bmcAsSat}, and theorem proving \cite{smt}. SAT was the first problem to be shown to be NP-complete \cite{cook-npcompleteness}, and there is believed to be no general procedure for solving arbitrary SAT instances efficiently. Nevertheless, modern solvers are able to routinely decide large SAT instances in practice, with different algorithms proving to be more successful than others on particular problem instances. For example, incomplete search methods such as WalkSAT \cite{walksat} and survey propagation \cite{surveyProp} are more effective at solving large, randomly generated formulas, while complete solvers leveraging \textit{conflict-driven clause learning} (CDCL) \cite{CDCL} fare better on large structured SAT formulas that commonly arise in industrial settings.

% develop these solvers relies on realistic SAT benchmarks
Understanding, developing and evaluating modern SAT solvers relies heavily on extensive empirical testing on a suite of benchmark SAT formulas. Unfortunately, in many domains, availability of benchmarks is still limited. While this situation has improved over the years, new and interesting benchmarks --- both real and synthetic --- are still in demand and highly welcomed by the SAT community. Developing expressive generators of structured SAT formulas is important, as it would provide for a richer set of evaluation benchmarks, which would in turn allow for the development of better and faster SAT solvers. Indeed, the problem of pseudo-industrial SAT formula generation has been identified as one of the ten key challenges in propositional reasoning and search \cite{10challenges}. 

One promising approach for tackling this challenge is to represent SAT formulas as graphs, thus recasting the original problem as a graph generation task. Specifically, every SAT formula can be converted into a corresponding bipartite graph, known as its literal-clause graph (LCG), via a bijective mapping. 
% However, generating LCGs is challenging, as they have to obey the hard bipartite constraint. 
Prior work in pseudo-industrial SAT instance generation has relied on hand-crafted algorithms \cite{mod_gen, loc_gen}, focusing on capturing one or two of the graph statistics exhibited by real-world SAT formulas \cite{scale-free,community-structure}. 
As researchers continue to uncover new and interesting characteristics of real-world SAT instances \cite{scale-free, eigen, community-structure, new_properties}, previous SAT generators might become invalid, and hand-crafting new models that simultaneously capture all the pertinent properties becomes increasingly difficult. On the other hand, recent work on deep generative models for graphs \cite{netgan, li2018learning, you2018graph, you2018graphrnn} has demonstrated their ability to capture \emph{many} of the essential features of real-world graphs such as social networks and citation networks, as well as graphs arising in biology and chemistry. However, these models do not enforce bipartiteness and therefore cannot be directly employed in our setting. While it might be possible to post-process these generated graphs, such a solution would be ad hoc, computationally expensive and would fail to exploit the unique structure of bipartite graphs.

%%%%%%%%%%%%% old version for archive purpose
% One promising direction to resolve the above challenge would be to represent industrial SAT formulas with graph representations.
% \jiaxuan{Dicussing that both LCGs LIGs are natural choices, but each with individual challenges}.
% such as Literal-Incidence Graphs (LIGs), which exhibit unique graph statistics, such as modularity and scale-free structures\cite{community-structure,scale-free}, and LCGs which can be projected to SAT formulas via a bijective function. 
% % \jiaxuan{existing doesn't work well}
% Existing work focused on adapting classic random graph generators, focusing on fitting \emph{one} of the graph statics to generate LIGs \cite{mod_gen,loc_gen}. While these models enable theoretical analysis, they are heavily hand-crafted and cannot capture \emph{many} of the essential characteristics exhibited by industrial SAT formulas. Moreover, as researchers continue to uncover unique characteristics of industrial SAT instances, previous SAT generators might become invalid, and designing new models that capture all known characteristics is becoming increasingly difficult.
%%%%%%%%%%%%% old version for archive purpose

\begin{figure}[t]
\centering
\includegraphics[width=\linewidth]{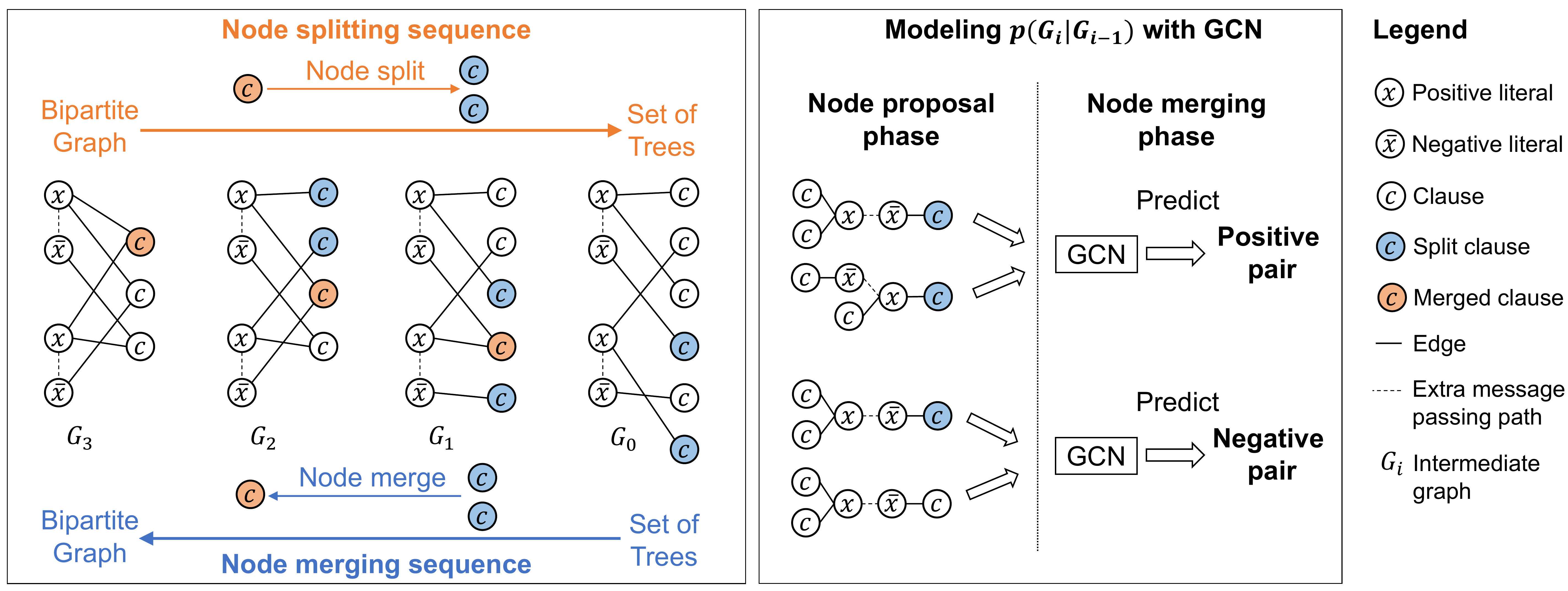} 
\caption{An overview of the proposed G2SAT model. 
\textbf{Top left}: A given bipartite graph can be decomposed into a set of disjoint trees by applying a sequence of node splitting operations. Orange node $c$ in graph $G_i$ is split into two blue $c$ nodes in graph $G_{i-1}$. Every time a node is split, one more node appears in the right partition. 
\textbf{Right}: We use node pairs gathered from such a sequence of node splitting operations to train a GCN-based classifier that predicts whether a pair of $c$ nodes should be merged. 
\textbf{Bottom left}: Given such a classifier, G2SAT generates a bipartite graph by starting with a set of trees $G_0$ and applying a sequence of node merging operations, where two blue nodes in graph $G_{i-1}$ get merged in graph $G_{i}$. G2SAT uses the GCN-based classifier that captures the bipartite graph structure to sequentially decide which nodes to merge from a set of candidates. Best viewed in color.
}
\vspace{-3mm}
\label{fig:framework}
\end{figure}

% Here we focus on 
% \jiaxuan{Here we show that the deep graph generative model is a universal solution for SAT generation.}
In this paper, we present G2SAT, the first deep generative model that \emph{learns} to generate SAT formulas based on a given set of input formulas. 
% The core of G2SAT is a novel deep generative model specialized in bipartite graph generation.
We use LCGs to represent SAT formulas, and formulate the task of SAT formula generation as a bipartite graph generation problem. Our key insight is that any bipartite graph can be generated by starting with a set of trees, and then applying a sequence of {\em node merging} operations over the nodes from one of the two partitions. As we merge nodes, trees are also merged, and complex bipartite structures begin to appear (Figure~\ref{fig:framework}, left). In this manner, a set of input bipartite graphs (SAT formulas) can be characterized by a distribution over the sequence of node merging operations. Assuming we can capture/learn the distribution over the pairs of nodes to merge, we can start with a set of trees and then keep merging nodes in order to generate realistic bipartite graphs (i.e., realistic SAT formulas). G2SAT models this iterative node merging process in an auto-regressive manner, where a node merging operation is viewed as a sample from the underlying conditional distribution that is parameterized by a Graph Convolutional Neural Network (GCN) \cite{hamilton2017inductive,hamilton2017representation,kipf2016semi,you2019position}, and the same GCN is shared across all the steps of the generation process. 

% A challenge we have to resolve is that the generative process is sequential, while SAT formulas are static. 
% To resolve the challenge and generate training data for our generator, we proceed as follows (Figure~\ref{fig:framework}).
This formulation raises the following question: how do we devise a sequential generative process when we are only given a static input SAT formula? In other words, how do we generate training data for our generator?
% A challenge we have to resolve is how to generate training data for our generator, which corresponds to inferring the sequential generative process from the static input SAT formulas.
We resolve this challenge as follows (Figure~\ref{fig:framework}). We define {\em node splitting} as the inverse operation to node merging. We apply this node splitting operation to a given input bipartite graph (a real-world SAT formula) and decompose it into a set of trees. We then reverse the splitting sequence, so that we start with a set of trees %\raghu{The term ``clause'' hasn't been defined yet; perhaps drop the following parenthetical?} (which describe SAT clause sizes) 
and learn from the sequence of node merging operations that recovers a realistic SAT formula. We train a GCN-based classifier that decides which two nodes to merge next, based on the structure of the graph generated so far.

At graph generation time, we initialize G2SAT with a set of trees and iteratively merge %sample \raghu{should this be ``merge'' instead of ``sample''?} random
node pairs based on the conditional distribution parameterized by the trained GCN model, until a user-specified stopping criterion is met. We utilize an efficient two-phase generation procedure: in the node proposal phase, candidate node pairs are randomly drawn, and in the node merging phase, the learned GCN model is applied to select the most likely node pair to merge. %\jiaxuan{G2SAT model explained}

Experiments demonstrate that G2SAT is able to generate formulas that closely resemble the input real-world SAT instances in many graph-theoretic properties such as modularity and the presence of scale-free structures, with $24\%$ lower relative error compared to state-of-the-art methods. 
More importantly, G2SAT generates formulas that exhibit the same hardness characteristics as real-world SAT formulas in the following sense: when the generated instances are solved using various SAT solvers, those solvers that are known to be effective on structured, real-world instances consistently outperform those solvers that are specialized in solving random SAT instances.
% exhibit the relative hardness of real-world SAT formulas. In particular, we apply different SAT solvers to formulas generated by G2SAT. We find that SAT solvers that specialize in real-world SAT instances consistently outperform solvers that specialize in random SAT instances on graphs generated by G2SAT. This suggests that G2SAT learns to generate SAT instances that are more similar to real-world instances than to random SAT instances. 
Moreover, our results suggest that we can use our synthetically generated formulas to more effectively tune the hyperparameters of SAT solvers, achieving an $18\%$ speedup in run time on unseen formulas, compared to tuning the solvers only on the formulas used for training.\footnote{Link to code and datasets: \url{http://snap.stanford.edu/g2sat/}}

\section{Preliminaries}
\label{sec:prelim}

%\raghu{I reordered the sections, so Preliminaries now comes before the Related Work section. Otherwise, the latter ends up using a bunch of terms that haven't been defined yet (LIG, VCG, etc.)}
% \jiaxuan{Shrink to 2 paragraphs, focus on the goal and related definitions}

\xhdr{Goal of generating SAT formulas}
Our goal is to design a SAT generator that, given a suite of SAT formulas, generates new SAT formulas with similar properties. Our aim is to capture not only graph theoretic properties, but also realistic SAT solver behavior. 
For example, if we train our G2SAT model on formulas from application domain $X$, then solvers that traditionally excel in solving problems in domain $X$, should also excel in solving synthetic G2SAT formulas (rather than, say, solvers that specialize in solving random SAT formulas).

%we can claim that the formulas we generate are realistic  on those formulas, solvers that traditionally excel in industrial formulas outperform solvers that specialize in random formulas.

\xhdr{SAT formulas and their graph representations}
A SAT formula $\phi$ is composed of Boolean variables $x_i$ connected by the logical operators \emph{and} ($\land$), \emph{or} ($\lor$), and \emph{not} ($\neg$). A formula is satisfiable if there exists an assignment of Boolean values to the variables such that the overall formula evaluates to true. 
In this paper, we are concerned with formulas in Conjunctive Normal Form (CNF)\footnote{Any SAT formula can be converted to an equisatisfiable CNF formula in linear time \cite{tseytin}.}, i.e., formulas expressed as conjunctions of disjunctions. Each disjunction is called a \textit{clause}, while a Boolean variable $x_i$ or its negation $\neg x_i$ is called a \textit{literal}.
% \footnote{Any SAT formula can be converted in linear time to an equisatisfiable CNF formula with linear growth in formula size \cite{tseytin}.}
For example, $(x_1 \lor x_2 \lor \neg x_3) \land (\neg x_1 \lor \neg x_2)$ is a CNF formula with two clauses that can be satisfied by assigning true to $x_1$ and false to $x_2$. 

Traditionally, researchers have studied four different graph representations for SAT formulas \cite{satgraph}: (1) \emph{Literal-Clause Graph (LCG)}: there is a node for each literal and each clause, with an edge denoting the occurrence of a literal in a clause. An LCG is bipartite and there exists a bijection between CNF formulas and LCGs. (2) \emph{Literal-Incidence Graph (LIG)}: there is a node for each literal and two literals have an edge if they co-occur in a clause. (3) \emph{Variable-Clause Graph (VCG)}: obtained by merging the positive and negative literals of the same variables in an LCG. (4) \emph{Variable-Incidence Graph (VIG)}: obtained by performing the same literal merging operation on the LIG. In this paper, we use LCGs to represent SAT formulas.

\xhdr{LCGs as bipartite graphs}
%LCGs are bipartite graphs with the vacuous-clause freedom constraint.
We represent a bipartite graph $G=(\mathcal{V}^{G}, \mathcal{E}^{G})$ by its node set $\mathcal{V}^{G}=\{v_1^{G},...,v_n^{G}\}$ and edge set $\mathcal{E}^{G}\subseteq\{(v_i^{G},v_j^{G})|v_i^{G},v_j^{G}\in \mathcal{V}^{G}\}$. In the rest of paper, we omit the superscript $G$ whenever it is possible to do so.
% there is no ambiguity on which graph we are referring to.
Nodes in a bipartite graph can be split into two disjoint partitions $\mathcal{V}_1$ and $\mathcal{V}_2$ such that $\mathcal{V}=\mathcal{V}_1 \cup \mathcal{V}_2$. Edges only exist between nodes in different partitions, i.e., $\mathcal{E}\subseteq\{(v_i,v_j)|v_i\in \mathcal{V}_1, v_j\in \mathcal{V}_2\}$. An LCG with $n$ literals and $m$ clauses has $\mathcal{V}_1=\{l_1,...,l_n\}$ and $\mathcal{V}_2=\{c_1,...,c_m\}$, where $\mathcal{V}_1$ and $\mathcal{V}_2$ are referred to as the literal partition and the clause partition, respectively. We may also write out $l_i$ as $x_i$ or $\neg x_i$ when specifying the literal sign is necessary.
% \jure{We always make it seem we have 1 node per varialbe, but here we have 2 nodes per variable. We have to explain why, otherwise people will be confused.}

%Vacuous-clause freedom constraint can then be expressed as, if $(x_i, c_i)\in \mathcal{E}$, then $(\neg x_i, c_i)\notin \mathcal{E}$. 
% \jure{There is a bug in the last sentence $(x_i, c_i)$ cannot be in and out of  $\mathcal{E}$.}

% \xhdr{Comparison of different graph representations}
\xhdr{Benefits of using LCGs to generate SAT formulas}
While we choose to work with LCGs because they are bijective to SAT formulas, the LIG is also a viable alternative. Unlike LCGs, there are no explicit constraints over LIGs, and thus, previously developed general deep graph generators could in principle be used. 
% Moreover, the size of a LIG is usually much smaller than that of the corresponding LCG, thus potentially leading to faster training and generation. 
However, the ease of generating LIGs is offset by the fact that key information is lost during the translation from the corresponding SAT formula. In particular, given a pair of literals, the LIG only encodes whether they co-occur in a clause but fails to capture how many times and in which clauses they co-occur. It can further be shown that an LIG corresponds to a number of SAT formulas that is at least exponential in the number of 3-cliques in the LIG. This ambiguity severely limits the usefulness of LIGs for SAT benchmark generation.

% Moreover, it can be be proved that the number of SAT formulas that an LIG could be mapped to is equal to the number of clique edge covers in the graph. There is no straightforward and convincing way to choose from such a large set of choices.

% \xhdr{G2SAT framework} 
% We explored both options by presenting two generators, LIG2SAT and LCG2SAT. LIG2SAT introduces a novel algorithm that efficiently maps a LIG to a SAT formula, leveraging knowledge of the structures of SAT formulas. On the other hand, LCG2SAT introduces a novel and efficient deep generative model for bipartite graphs.
% Both generators involve two steps. First, we use graph generative models to learn the graph representations of real-world formulas and generate similar graphs. We then recover SAT formulas from the generated graphs. 

\section{Related Work}
\label{sec:related}

\xhdr{SAT Generators}
Existing synthetic SAT generators are hand-crafted models that are typically designed to generate formulas that fit a particular graph statistic. The mainstream generators for pseudo-industrial SAT instances include the Community Attachment (CA) model \cite{mod_gen}, which generates formulas with a given VIG modularity, and the Popularity-Similarity (PS) model \cite{loc_gen}, which generates formulas with a specific VCG degree distribution. In addition, there are also generators for random $k$-SAT instances \cite{random-ksat} and crafted instances that come from translations of structured combinatorial problems, such as graph coloring, graph isomorphism, and Ramsey numbers \cite{cnfGen}. Currently, all SAT generators are hand-crafted and machine learning provides an exciting alternative.

\xhdr{Deep Graph Generators}
Existing deep generative models of graphs fall into two categories.
In the first class are models that focus on generating perturbations of a given graph, by direct decoding from computed node embeddings \cite{kipf2016variational} or latent variables \cite{grover2018graphite}, or learning 
% the sampling strategy based on 
the random walk distribution of a graph \cite{netgan}.
The second class comprises models that can learn to generate a graph by sequentially adding nodes and edges \cite{li2018learning, you2018graphrnn, you2018graph}. Domain specific generators for molecular graphs \cite{de2018molgan,jin2018junction} and 3D point cloud graphs \cite{valsesia2018learning} have also been developed. However, current deep generative models of graphs do not readily apply to SAT instance generation. Thus, we develop a novel bipartite graph generator that respects all the constraints imposed by graphical representations of SAT formulas and generates the formula graph via a sequence of node merging operations.

\xhdr{Deep learning for SAT}
NeuroSAT also represents SAT formulas as graphs and computes node embeddings using GCNs \cite{selsam2018learning}. However, NeuroSAT focuses on using the embeddings to solve SAT formulas, while we aim to generate SAT formulas. A preliminary version of the work presented in this paper appeared in \cite{satgen}, where existing graph generative models were used to learn the LIG of a SAT formula. However, extensive post-processing was required to extract a formula from a generated LIG, since an LIG is an ambiguous representation of a SAT formula. In this work, we develop a new deep graph generative model, that, unlike existing techniques, is able to directly learn the bijective graph representation of a SAT formula, and therefore better capture its characteristics.

\section{The G2SAT Framework}
\label{sec:g2sat}

% \section{Proposed Method}

% \jiaxuan{I'll expand this section to be 2 pages, while also create a figure illustate the model}

\subsection{G2SAT: Generating Bipartite Graphs by Iterative Node Merging Operations}
%\raghu{I think the notation in the following could be cleaned up. $p(G)$ is the true distribution and $p_{model}(G)$ is the learned distribution; but the latter never gets used after being defined (I think $p(G)$ is getting overloaded in what follows below). Also, something like $\hat{p}(G)$ may be less clunky than $p_{model}(G)$.}
As discussed in Section \ref{sec:prelim}, a SAT formula is uniquely represented by its LCG which is a bipartite graph. From the perspective of generative models, our primary objective is to learn a distribution %$\hat{p}(G)$ 
over bipartite graphs, based on a set of observed bipartite graphs $\mathbb{G}$ sampled from the data distribution $p(G)$. Each bipartite graph $G \in \mathbb{G}$ may have a different number of nodes and edges. Due to the complex dependency between nodes and edges, directly learning $p(G)$ is challenging. Therefore, we generate a graph via an $n$-step iterative process, $p(G) = \prod_{i=1}^{n}{p(G_i|G_1,...,G_{i-1})}$,
% \begin{equation}
%      p(G) = \prod_{i=1}^{n}{p(G_i|G_1,...,G_{i-1})}
% \end{equation}
where $G_i$ refers to an intermediate graph at step $i$ in the iterative generation process. 
% The key of a successful iterative graph generator is to instantiate the conditional distribution $p(G_i|G_1,...,G_{i-1})$.
Since we focus on generating static graphs, we assume that the order of the generation trajectory does not matter, as long as the same graph is generated. This assumption implies the following Markov property over the conditional distribution, $p(G_i|G_1,...,G_{i-1}) = p(G_i|G_{i-1})$.

The key to a successful iterative graph generative model is a proper instantiation of the conditional distribution $p(G_i|G_{i-1})$.
Existing approaches \cite{li2018learning, you2018graph, you2018graphrnn} often model $p(G_i|G_{i-1})$ as the distribution over the random addition of nodes or edges to $G_{i-1}$. While in theory this formulation allows the generation of any kind of graph, it cannot satisfy the hard partition constraint for bipartite graphs. In contrast, our proposed G2SAT has a simple generation process that is guaranteed to preserve the bipartite partition constraint, without the need for hand-crafted generation rules or post-processing procedures. The G2SAT framework relies on node splitting and merging operations, which are defined as follows.
\begin{definition}
The node splitting operation, when applied to node $v$, removes some edges between $v$ and its neighboring nodes, and then connects those edges to a new node $u$. 
The node merging operation, when performed over two nodes $u$ and $v$, removes all the edges between $v$ and its neighboring nodes, and then connects those edges to $u$.
Formally, $\NodeSplit(u, G)$ returns a tuple $(u,v,G')$, and $\NodeMerge(u,v,G)$ returns a tuple $(u,G')$.
\end{definition}
Note that according to this definition, a node merging operation can always be reversed by a node splitting operation. The core idea underlying G2SAT is then motivated by the following observation.
\begin{observation}
\label{ob:1}
A bipartite graph can always be transformed into a set of trees by a sequence of node splitting operations over the nodes in one of the partitions. 
\end{observation}
The proof of this claim follows from the fact that the node splitting operation strictly reduces a node's degree. Therefore, repeatedly applying node splitting to all the nodes in a partition will ultimately reduce the degree of all those nodes to $1$, producing a set of trees (Figure \ref{fig:framework}, Left). This observation implies that a bipartite graph can always be generated via \emph{a sequence of node merging operations}. In G2SAT, we always merge clause nodes in the clause partition $\mathcal{V}_2^{G_{i-1}}$ for a given graph $G_{i-1}$.
% , for example, merging $u$, $v$ in graph $G_{i-1}$ can be written as $G_i = \NodeMerge(u, v, G_{i-1})$. 
We then design the following instantiation of $p(G_i|G_{i-1})$,
\begin{equation}
\label{eq:conditional}
    p(G_i|G_{i-1}) = p(\NodeMerge(u, v, G_{i-1})| G_{i-1}) = \Multimonial(\mb{h}_u^T\mb{h}_v/Z|\forall u,v \in \mathcal{V}_2^{G_{i-1}})
\end{equation}
where $\mb{h}_u$ and $\mb{h}_v$ are the embeddings for nodes $u$ and $v$, and $Z$ is the normalizing constant that ensures that the distribution $\Multimonial(\cdot)$ is valid. We aim for embeddings $\mb{h}_u$ that capture the multi-hop neighborhood structure of a node $u$ and that can be computed from a single trainable model. Further, this model needs to be capable of generalizing across different generation stages and different graphs. Therefore, we use the GraphSAGE framework \cite{hamilton2017inductive} to compute node embeddings, which is a variant of GCNs that has been shown to have strong inductive learning capabilities across different graphs. Specifically, the $l$-th layer of GraphSAGE can be written as %\raghu{if the following equation is not subsequently referenced, there's no need to label it.}
% \begin{equation}
%     H^{(l+1)} = \text{ReLU}(\{\tilde{D}^{-\frac{1}{2}}\tilde{E}\tilde{D}^{-\frac{1}{2}} H^{(l)} W^{(l)}\}
% \end{equation}
\begin{equation*}
\begin{aligned}
    & \mathbf{n}_u^{(l)} = \textsc{AGG}(\textsc{ReLU}(\mathbf{Q}^{(l)}\mathbf{h}_v^{(l)}+\mathbf{q}^{(l)}|v\in N(u))) \\
    & \mathbf{h}_u^{(l+1)} = \textsc{ReLU}(\mathbf{W}^{(l)}\textsc{concat}(\mathbf{h}_u^{(l)}, \mathbf{n}_u^{(l)}))
\end{aligned}
\end{equation*}
where $\mathbf{h}_u^{(l)}$ is the $l$-th layer node embedding for node $u$, $N(u)$ is the local neighborhood of $u$, $\textsc{AGG}(\cdot)$ is an aggregation function such as mean pooling, and $\mathbf{Q}^{(l)}, \mathbf{q}^{(l)}, \mathbf{W}^{(l)}$ are trainable parameters.
% Given graph $G_{i-1}$ in the last step, LCG2SAT first samples a pair of nodes based on Equation \ref{eq:conditional}, then conduct node merging to get $G_i$ for the next step.
The input node features are length-3 one-hot vectors, which are used to represent the three node types in LCGs, i.e., positive literals, negative literals and clauses. In addition, since each literal and its negation are closely related, we add an additional message passing path between them. 

\subsection{Scalable G2SAT with Two-phase Generation Scheme}
\label{sc:two-phase}
% \jiaxuan{I'm looking into literature to find some justification / better explanation}
% \jiaxuan{Got the justification, it's Noise contrastive sampling. Reformulating this part}
% The computation of normalizing constant $Z$ in Equation \ref{eq:conditional} is 
LCGs can easily have tens of thousands of nodes; thus, there are millions of candidate node pairs that could be merged. This makes the computation of the normalizing constant $Z$ in Equation \ref{eq:conditional} infeasible. 
To avoid this issue, we design a two-phase scheme to instantiate Equation \ref{eq:conditional}, which includes a node proposal phase and a node merging phase (Figure \ref{fig:framework}, right). Intuitively, the idea is to begin with a fixed oracle that proposes random candidate node pairs. Then, a model only needs to decide if the proposed node pair should be merged or not, which is an easier learning task compared to selecting from among millions of candidate options. Instead of directly learning and sampling from $p(G_i|G_{i-1})$, we introduce additional random variables $u$ and $v$ to represent random nodes, and then learn the joint distribution $p(G_i,u,v|G_{i-1})= p(u,v|G_{i-1}) p(G_i|G_{i-1},u,v)$. Here, $p(u,v|G_{i-1})$ corresponds to the node proposal phase and $p(G_i|G_{i-1},u,v)$ models the node merging phase.

In theory, $p(u,v|G_{i-1})$ can be any distribution as long as it has non-empty support. Since LCGs are inherently static graphs, there is little prior knowledge or additional information on how this iterative generation process should proceed. Therefore, we implement the node proposal phase such that a random node pair is sampled from all candidate clause nodes uniformly at random. Then, in the node merging phase, instead of computing the dot product between all possible node pairs, the model only needs to compute the dot product between the sampled node pairs. Specifically, we have %\raghu{which equation do you want to attach the label to below?}
\begin{align}
    p(G_i, u, v|G_{i-1}) = &~p(u,v | G_{i-1})p(\NodeMerge(u, v, G_{i-1})| G_{i-1}, u, v)\nonumber \\
     = & \Uniform(\{(u,v)|\forall u,v \in \mathcal{V}_2^{G_{i-1}}\}) \Bernoulli(\sigma(\mb{h}_u^T\mb{h}_v)|u,v)\label{eq:conditional_two_phase}
\end{align}
where $\Uniform$ is the discrete uniform distribution and $\sigma(\cdot)$ is the sigmoid function.

% $u$, $v$ are proposed node pairs,

\subsection{G2SAT at Training Time}

% How we prepare training data.
% The challenge we have to resolve is that the generative process is sequential, while SAT formulas are static. 
% To resolve the challenge and generate training data for our G2SAT generator we proceed as follows (Figure~\ref{fig:framework}).

The two-phase generation scheme described in Section \ref{sc:two-phase} transforms the bipartite graph generation task into a binary classification task. We train the classifier to minimize the following binary cross entropy loss:
\begin{equation}
\label{eq:obj}
    \mathcal{L} = -\mathbb{E}_{u,v\sim p_{pos}}[\log(\sigma(\mathbf{h}_u^T \mathbf{h}_v))] - \mathbb{E}_{u,v\sim p_{neg}}[\log(1-\sigma(\mathbf{h}_u^T \mathbf{h}_v))]
\end{equation}
% \begin{wraptable}{r}{.56\linewidth}
% \begin{minipage}{0.49\textwidth}
% \begin{algorithm}[H]
% \caption{G2SAT at training time}
% \label{alg:train}
% \begin{algorithmic}
% \STATE {\bfseries Input:} Bipartite graphs $\mathcal{G}$, repeat time $r$
% \STATE {\bfseries Output:} Graph templates $\mathcal{T}$
% \STATE $\mathcal{D} \leftarrow \varnothing$, $\mathcal{T} \leftarrow \varnothing$
% \FOR{$k=1,\dots,r$}
% \STATE $G \sim \mathcal{G}$, $n \leftarrow |\mathcal{E}^{G}|-|\mathcal{V}^{G}_2|$, $G_n \leftarrow G$
% \FOR{$i=n,\dots,1$} 
%     \STATE $s \sim \{u|u \in \mathcal{V}_2^{G_i}, \Degree(u)>1\}$
%     \STATE $(u^{+}, v^{+},G_{i-1}) \leftarrow \NodeSplit(s, G_i)$
%     \STATE $v^{-} \sim \mathcal{V}_2^{G_i}\setminus\{u^{+}, v^{+}\}$
%     \STATE $\mathcal{D} \leftarrow \mathcal{D} \cup \{(u^{+},v^{+},v^{-}, G_i)\}$
% \ENDFOR   
% \STATE $\mathcal{T} \leftarrow \mathcal{T} \cup \{(G_0, n)\}$
% \ENDFOR
% \STATE Train G2SAT with $\mathcal{D}$ to minimize Eq. \ref{eq:obj}
% %  $\mb{h}_v \leftarrow \mb{h}_v^L \in \mathbb{R}^{s_l}$  \\
% \end{algorithmic}
% \end{algorithm}
% \end{minipage}
% \end{wraptable}
where $p_{pos}$ and $p_{neg}$ are the distributions over positive and negative training examples (i.e. node pairs). We say a node pair is a positive example %\raghu{By positive and negative, are you talking about positive and negative training examples for the classifier? I think this needs to be rephrased.} 
if the node pair should be merged according to the training set. %, and vice versa \raghu{Not sure what the ``vice versa'' refers to. I think this sentence needs to rephrased}. 
To acquire the necessary training data from input bipartite graphs, we develop a procedure that is described in Algorithm \ref{alg:train}.
Given an input bipartite graph $G$, we apply the node splitting operation to the graph for $n=|\mathcal{E}^{G}|-|\mathcal{V}^{G}_2|$ steps, which guarantees that the input graph will be decomposed into a set of trees. Within each step, a random node $s$ in partition $\mathcal{V}_2^{G_i}$ with degree greater than $1$ is chosen for splitting, and a random subset of edges that connect to $s$ is chosen to connect to a new node. After the split operation, we obtain an updated graph $G_{i-1}$, as well as the split nodes $u^{+}$ and $v^{+}$, which are treated as a positive training example. %\raghu{``training example'' instead of ``node pair''?}. 
Then, another node $v^{-}$, that is distinct from $u^{+}$ and $v^{+}$, is randomly chosen from the nodes in $\mathcal{V}_2^{G_{i-1}}$, and $(u^{+}, v^{-})$ are viewed as a negative training example. The data tuple $(u^{+},v^{+},v^{-}, G_{i-1})$ is saved in the dataset $\mathcal{D}$. We also save the step count $n$ and the graph $G_0$ as ``graph templates'', which are later used to initialize G2SAT at inference time. 
The procedure is repeated $r$ times until the desired number of data points are gathered. Finally, G2SAT is trained with the dataset $\mathcal{D}$ to minimize the objective listed in Equation \ref{eq:obj}. %\raghu{Can we spare a little more space around the algorithms? They're a little too tight right now.}

\begin{minipage}{0.49\textwidth}
\vspace{-2mm}
\begin{algorithm}[H]
\caption{G2SAT at training time}
\label{alg:train}
\begin{algorithmic}
\STATE {\bfseries Input:} Bipartite graphs $\mathcal{G}$, number of repetitions $r$
\STATE {\bfseries Output:} Graph templates $\mathcal{T}$
\STATE $\mathcal{D} \leftarrow \varnothing$, $\mathcal{T} \leftarrow \varnothing$
\FOR{$k=1,\dots,r$}
\STATE $G \sim \mathcal{G}$, $n \leftarrow |\mathcal{E}^{G}|-|\mathcal{V}^{G}_2|$, $G_n \leftarrow G$
\FOR{$i=n,\dots,1$} 
    \STATE $s \sim \{u|u \in \mathcal{V}_2^{G_i}, \Degree(u)>1\}$
    \STATE $(u^{+}, v^{+},G_{i-1}) \leftarrow \NodeSplit(s, G_i)$
    \STATE $v^{-} \sim \mathcal{V}_2^{G_{i-1}}\setminus\{u^{+}, v^{+}\}$
    \STATE $\mathcal{D} \leftarrow \mathcal{D} \cup \{(u^{+},v^{+},v^{-}, G_{i-1})\}$
\ENDFOR   
\STATE $\mathcal{T} \leftarrow \mathcal{T} \cup \{(G_0, n)\}$
\ENDFOR
\STATE Train G2SAT with $\mathcal{D}$ to minimize Eq. \ref{eq:obj}
%  $\mb{h}_v \leftarrow \mb{h}_v^L \in \mathbb{R}^{s_l}$  \\
\end{algorithmic}
\end{algorithm}
\vspace{-4mm}
\end{minipage}
\hfill
\begin{minipage}{0.49\textwidth}
\vspace{-1mm}
\begin{algorithm}[H]
\caption{G2SAT at inference time}
\label{alg:inference}
\begin{algorithmic}
\STATE {\bfseries Input:} Graph templates $\mathcal{T}$, number of output graphs $r$, number of proposed node pairs $o$
\STATE {\bfseries Output:} Generated bipartite graphs $\mathcal{G}$
% \STATE $\mathcal{G} \leftarrow \varnothing$
\FOR{$k=1,\dots,r$}
\STATE $(G_0,n) \sim \mathcal{T}$
\FOR{$i=0,\dots,n-1$}
    \STATE $\mathcal{P} \leftarrow \varnothing$
    \FOR{$j=1,\dots,o$}
    \STATE $u \sim \mathcal{V}_2^{G_i}$, 
    $v \sim \{s|s \in \mathcal{V}_2^{G_i}, (s,x) \notin \mathcal{E}^{G_i}, (s,\neg x) \notin \mathcal{E}^{G_i}, \forall x \in N(u)\}$
    \STATE $\mathcal{P} = \mathcal{P} \cup \{(u,v)\}$
    \ENDFOR
    % \STATE $\{(\mb{h}_u, \mb{h}_v)\} \leftarrow \GCN(\mathcal{P},G_i)$
    \STATE $(u^{+}, v^{+}) \leftarrow \argmax\{\mb{h}_u^T\mb{h}_v|(u,v)\in \mathcal{P}, \mb{h}_u=\GCN(u), \mb{h}_v=\GCN(v)\}$
    \STATE $G_{i+1} \leftarrow \NodeMerge(u^{+}, v^{+},G_i)$
\ENDFOR   
\STATE $\mathcal{G} = \mathcal{G} \cup \{G_n\}$
\ENDFOR
% \STATE Train G2SAT with positive and negative pairs 
\end{algorithmic}
\end{algorithm}
\vspace{-10mm}
\end{minipage}

\subsection{G2SAT at Inference Time}
% How we infer initial state, sample from conditional distribution.
A trained G2SAT model can be used to generate graphs. We summarize the procedure in Algorithm \ref{alg:inference}.
At graph generation time, we first initialize G2SAT with a graph template sampled from $\mathcal{T}$ gathered at training time, which specifies the initial graph $G_0$ and the number of generation steps $n$. Note that G2SAT can take bipartite graphs with arbitrary size as input and iterate for a variable number of steps. The reason we specify the initial state and the number of steps is to control the behavior of G2SAT and simplify the experiment setting.
% \begin{wraptable}{r}{.56\linewidth}
% \begin{minipage}{0.49\textwidth}
% \begin{algorithm}[H]
% \caption{G2SAT at inference time}
% \label{alg:inference}
% \begin{algorithmic}
% \STATE {\bfseries Input:} Graph templates $\mathcal{T}$, number of output graph $r$, number of proposed node pairs $o$
% \STATE {\bfseries Output:} Generated bipartite graphs $\mathcal{G}$
% % \STATE $\mathcal{G} \leftarrow \varnothing$
% \FOR{$k=1,\dots,r$}
% \STATE $(G_0,n) \sim \mathcal{T}$
% \FOR{$i=0,\dots,n-1$}
%     \STATE $\mathcal{P} \leftarrow \varnothing$
%     \FOR{$j=1,\dots,o$}
%     \STATE $u \sim \mathcal{V}_2^{G_i}$, 
%     $v \sim \{s|s \in \mathcal{V}_2^{G_i}, (s,x) \notin \mathcal{E}^{G_i}, (s,\neg x) \notin \mathcal{E}^{G_i}, \forall x \in N(u)\}$
%     \STATE $\mathcal{P} = \mathcal{P} \cup \{(u,v)\}$
%     \ENDFOR
%     % \STATE $\{(\mb{h}_u, \mb{h}_v)\} \leftarrow \GCN(\mathcal{P},G_i)$
%     \STATE $(u^{+}, v^{+}) \leftarrow \argmax\{\mb{h}_u^T\mb{h}_v|(u,v)\in \mathcal{P}, \mb{h}_u=\GCN(u), \mb{h}_v=\GCN(v)\}$
%     \STATE $G_{i+1} \leftarrow \NodeMerge(u^{+}, v^{+},G_i)$
% \ENDFOR   
% \STATE $\mathcal{G} = \mathcal{G} \cup \{G_n\}$
% \ENDFOR
% % \STATE Train G2SAT with positive and negative pairs 
% \end{algorithmic}
% \end{algorithm}
% \end{minipage}
% \end{wraptable}

At each generation step, we use the two-phase generation scheme described in Section \ref{sc:two-phase}. In the node proposal phase, we additionally make sure that the sampled node pair does not correspond to a vacuous clause, i.e., if $u, v$ is a valid node pair, then $\forall x \in N(u)$, we ensure that $(v,x) \notin \mathcal{E}^{G_i}$ and $(v,\neg x) \notin \mathcal{E}^{G_i}$.
We parallelize the algorithm by sampling $o$ random node pair proposals at once and feeding them to the node merging phase.
In the node merging phase, although following Equation \ref{eq:conditional_two_phase} would allow us to sample from the true distribution, we find in practice that it usually requires sampling a large number of candidate nodes pairs until a positive node pair is predicted by the GCN model.
Therefore, we use a greedy algorithm that selects the most likely node pair to be merged among the $o$ proposed node pairs and merge those nodes. Admittedly, this biases the generator away from the true data distribution. However, our experiments reveal that the synthesized graphs are nonetheless reasonable. After $n$ steps, the generated graph $G_n$ is saved as an output.

% and then iteratively sample random node pairs from the conditional distribution parametrized by the trained GCN model until user-specified stopping criteria are met. We develop an efficient two-phase generation procedure, where in the node proposal phase candidate node pairs are randomly drawn, then in the node merging phase G2SAT applies the learned GCN model to select the most likely node pair to be merged and merges the nodes.

\section{Experiments}
\label{sec:expts}
% \jiaxuan{In general, we just summarize key results, put other things to appendix.}

% In this section, we discuss in detail the experiments we performed to evaluate \satgen.

\subsection{Dataset and Evaluation}

\xhdr{Dataset}
We use 10 small real-world SAT formulas from the SATLIB benchmark library~\cite{satlib} and past SAT competitions.\footnote{\url{http://www.satcompetition.org/}} The two data sources contain SAT formulas generated from a variety of application domains, such as bounded model checking, planning, and cryptography. We use the standard SatElite preprocessor \cite{satelite} to remove duplicate clauses and perform polynomial-time simplifications (for example, unit propagation). The preprocessed formulas contain 82 to 1122 variables and 327 to 4555 clauses.

We evaluate if the generated SAT formulas preserve the properties of the input training SAT formulas, as measured by graph statistics and SAT solver performance. We then investigate whether the generated SAT formulas can indeed help in designing better domain-specific SAT solvers.
% To evaluate the efficacy of G2SAT, we compare the graph statistics between the formulas generated by G2SAT and formulas in the training set. 
% In addition, we perform a "mini SAT competition" on the formulas generated by G2SAT. In particular, we investigate whether on the formulas generated by G2SAT, SAT solvers that traditionally excel in real-world formulas outperform solvers that specialize in random formulas. 
% We take this as an indicator of whether G2SAT is able to capture the structures of real-world formulas. Finally, we used the synthetic formulas to tune the hyper-parameters of a SAT solver, and measure whether the optimal configuration over the synthetic formulas could translate to unseen benchmarks that have similar structures as the training set.

\xhdr{Graph statistics} We focus on the graph statistics studied previously in the SAT literature \cite{scale-free,community-structure}. In particular, we consider the VIG, VCG and LCG representations of SAT formulas.
We measure the modularity \cite{newman2006modularity} (in VIG, VCG, LCG), average clustering coefficient \cite{newman2001clustering} (in VIG) and the scale-free structure parameters as measured by variable $\alpha_v$ and clause $\alpha_c$ \cite{scale-free, max-likely} (in VCG). 
% To measure whether a formula has scale-free structures, we check whether the clause degrees and the variable degrees in the VCG respectively follow a power-law distribution. In other words, we verify whether there exist $\alpha_v$ and $\alpha_c$, such that the expected number of variables with degree $k$ in a VCG, $f_v(k)$, is approximately  $ck^{-\alpha_v}$ and the expected number of clauses with length $k$ in a VCG,  $f_c(k)$, is approximately $ck^{-\alpha_c}$ (where $c$ is a normalizing constant). 

% To measure whether a formula has scale-free structures, we check whether the clause degrees and the variable degrees in the VCG respectively follow a power-law distribution. In other words, we verify whether there exist $\alpha_v$ and $\alpha_c$, such that the expected number of variables with degree $k$ in a VCG, $f_v(k)$, is approximately  $ck^{-\alpha_v}$ and the expected number of clauses with length $k$ in a VCG,  $f_c(k)$, is approximately $ck^{-\alpha_c}$ (where $c$ is a normalizing constant). 

\xhdr{SAT solver performance} 
We report the relative SAT solver performance, i.e., given $k$ SAT solvers, we rank them based on their performance over the SAT formulas used for training and the generated SAT formulas, and evaluate how well the two rankings align. 
%Another option is to measure the absolute hardness of the formulas. %However, while it is not hard toas evidenced by extensive previous work, it is easy to increase a SAT formula's hardness regardless of the SAT solvers being used \cite{hardsat1, hardsat, trianglefree}. 
Previous research has shown that SAT instances can be made hard using various post-processing approaches \cite{trianglefree, hardsat1, hardsat}. Therefore, the absolute hardness of the formulas is not a good indicator of how realistic the formulas are. On the other hand, it is not trivial for a post-processing procedure to precisely manipulate the relative performance of a set of SAT solvers.
%However, previous research have shown that SAT instances could be made hard in various ways \cite{hardsat1, hardsat, trianglefree}. Therefore, the absolute hardness of the formulas is not a good indicator of how realistic the formulas are. On the other hand, it is not trivial to design an approach to precisely manipulating the performance ranking over a set of SAT solvers.
Therefore, we report the relative solver performance for a fairer comparison. We took the three best performing solvers from both the application track and the random track of the 2018 SAT competition \cite{sat2018}, which are denoted as $I_1, I_2, I_3$, and $R_1, R_2, R_3$ respectively.\footnote{The solvers are, in order, MapleLCMDistChronoBT, Maple\_LCM\_Scavel\_fix2, Maple\_CM, Sparrow2Riss-2018, gluHack, glucose-3.0\_PADC\_10\_NoDRUP \cite{sat2018}.}
%\raghu{Would be good to explicitly identify the solvers by name as well, perhaps in the appendix if there is going to be one.}. 
Our experiments confirm that solvers that are tailored to real-world SAT formulas ($I_1, I_2, I_3$) indeed outperform solvers that focus on random SAT formulas ($R_1, R_2, R_3$), over the training formulas. Therefore, we measure if on the generated formulas, the solvers $I$ similarly outperform the solvers $R$, as measured by ranking accuracy.
% One might argue that the absolute hardness of a formulas is a better metric for evaluating generated formulas. 
% claim that  being able to preserve relative solver behavior is a better indicator of the fact that a generator learns deeper properties of the training formulas.
All the run time performances are measured by wall clock time under carefully controlled experimental settings.% (detailed in the Appendix). \raghu{I don't see an Appendix.}
% \jiaxuan{Discuss why we don't compare exact run time here}

% specializing in random formulas consistently under-performing solvers specializing in real-world formulas, as evidence that the formulas \name~ generates are realistic.

% We took winners of the main (application) track and the random track of the 2018 SAT competition, and used them to solve the formulas generated by G2SAT. We compared the solver ranking on the generated formulas with the results of the 2018 SAT competition. In essence, we were measuring how well our generated formulas are able to classify solvers into those that excel in real formulas and those that excel in uniformly random formulas, using the 2018 SAT Competition results as the ground truth. We take the fact that solvers specializing in random formulas consistently under-performing solvers specializing in real-world formulas, as evidence that the formulas \name~ generates are realistic.

%The reason that we chose to judge the realistic-ness of the generated formulas based on relative run time, as opposed to absolute time, is that it is unrealistic to expect the generated formulas to have similar hardness as the formulas in the training set. 

\xhdr{Application: Developing better SAT solvers}
Finally, we consider the scenario where people wish to use the synthetic formulas for developing better SAT solvers. 
%Our experiment is for proof-of-concept purpose.
Specifically, we use either the 10 training SAT formulas or the generated SAT formulas to guide the hyperparameter selection of a popular SAT solver called Glucose \cite{glucose}. %\raghu{Why the Minisat citation?}. 
We conduct a grid search over two of its hyperparameters --- the variable decay $v_d$, that influences the ordering of the variables in the search tree, and the clause decay $c_d$, that influences which learned clauses are to be removed \cite{Minisat}. We sweep over the set $\{0.75, 0.85, 0.95\}$ for $v_d$, and the set $\{0.7, 0.8, 0.9, 0.99, 0.999\}$ for $c_d$. We measure the run time of the SAT solvers using the optimal hyperparameters found by grid search, over 22 real-world SAT formulas unobserved by any of the models. Since the number of training SAT formulas is limited, we expect that using the abundant generated SAT formulas will lead to better hyperparameter choices.
% , and measured whether the optimal configuration for the synthetic formulas improves the performance of Glucose compared with the optimal configuration for the training set.

% The former influences the variable ordering of the search tree, while the latter influences which learned clauses are to be removed \cite{Minisat}

\subsection{Models}
% \jiaxuan{Shrink}
We compare \name~with two state-of-the-art generators for real-world SAT formulas. Both generators are prescribed models designed to match a specific graph property. To properly generate formulas using these baselines, we set their arguments to match the corresponding statistics in the training set. We generate 200 formulas each using \name~and the baseline models.

\xhdr{G2SAT}
We implement G2SAT with a 3-layer GraphSAGE model using mean pooling and ReLU activation \cite{hamilton2017inductive} with hidden and output embedding size of 32. We use the Adam optimizer \cite{adam} with a learning rate of $0.001$ to train the model until the validation accuracy plateaus. 

\xhdr{Community Attachment (CA)}
The CA model generates formulas to fit a desired VIG modularity $Q$ \cite{mod_gen}. 
% The model takes in five inputs $n, m, k, c, Q$, where $n$ is the number of variables, $m$ the number of clauses, $k$ the length of each clause, $c$ the size of a partition of the VIG, and $Q$ is the desired VIG modularity. 
The output of the algorithm is a SAT formula with $n$ variables and $m$ clauses, each of length $k$, such that the optimal modularity for any $c$-partition of the VIG of the formula is approximately $Q$.

% The model takes in five inputs $n, m, k, c, Q$, where $n$ is the number of variables, $m$ the number of clauses, $k$ the length of each clause, $c$ the size of a partition of the VIG, and $Q$ is the desired VIG modularity. The output of the algorithm is a SAT formula with $n$ variables and $m$ clauses, each of length $k$, such that the optimal modularity for any $c$-partition of the VIG of the formula is approximately $Q$.

\xhdr{Popularity-Similarity (PS)}
The PS model generates formulas to fit desired $\alpha_v$ and $\alpha_c$ \cite{loc_gen}. The model accepts a temperature parameter $T$ that trades off the modularity and the $(\alpha_v, \alpha_c)$ measures of the generated formulas. We run PS with two temperature settings, $T=0$ and $T=1.5$.

% The PS model generates formulas to fit desired $\alpha_v$ and $\alpha_c$ \cite{loc_gen}. 
% In addition, the formulas generated by PS are guaranteed to have high modularity. 
% The model takes in seven inputs $n, m, k, K, \alpha_v, \alpha_c, T$, where $n$, $m$ are the same with CA, $k$ the minimum clause length, $K$ the average clause length, and $T$ a hyper-parameter that decides the trade-off between modularity and $\alpha_v, \alpha_c$. We use two versions of PS, with $T=0$ and $T=1.5$.

\subsection{Results}
\begin{table}[t]
\setlength\tabcolsep{2pt}
\centering		
\caption{Graph statistics of generated formulas (mean $\pm$ std. (relative error to training formulas)).} 
\label{table:stats}
\resizebox{1\columnwidth}{!}{ \renewcommand{\arraystretch}{1.1}
\hspace*{-0.2cm}\begin{tabular}{@{}clllllll@{}}\cmidrule[\heavyrulewidth]{2-8}
& \multirow{3}{*}{\vspace*{8pt}Method}&\multicolumn{2}{c}{VIG}&\multicolumn{3}{c}{VCG}&\multicolumn{1}{c}{LCG}\\
\cmidrule(lr){3-4} \cmidrule(lr){5-7} \cmidrule(lr){8-8}
& & \multicolumn{1}{c}{Clustering}& \multicolumn{1}{c}{Modularity} & \multicolumn{1}{c}{Variable $\alpha_v$} & \multicolumn{1}{c}{Clause $\alpha_c$} & \multicolumn{1}{c}{Modularity} & \multicolumn{1}{c}{Modularity}
\\ \cmidrule{2-8}
& Training & 0.50$\pm$0.07       & 0.58$\pm$0.09       & 3.57$\pm$1.08       & 4.53$\pm$1.09       & 0.74$\pm$0.06             & 0.63$\pm$0.05       \\ \cmidrule{2-8}
& CA           & 0.33$\pm$0.08(34\%) & 0.48$\pm$0.10(17\%) & 6.30$\pm$1.53(76\%) & N/A                    & 0.65$\pm$0.08(12\%)                  & 0.53$\pm$0.05(16\%) \\
& PS(T=0)     & 0.82$\pm$0.04(64\%) & 0.72$\pm$0.13(24\%) & 3.25$\pm$0.89(9\%)  & \textbf{4.70$\pm$1.59(4\%)}  & 0.86$\pm$0.05(16\%)   & \textbf{0.64$\pm$0.05(2\%)}  \\
& PS(T=1.5)   & 0.30$\pm$0.10(40\%) & 0.14$\pm$0.03(76\%) & 4.19$\pm$1.10(17\%) & 6.86$\pm$1.65(51\%) & 0.40$\pm$0.05(46\%)   & 0.41$\pm$0.05(35\%) \\
& G2SAT        & \textbf{0.41$\pm$0.09(18\%)} & \textbf{0.54$\pm$0.11(7\%)}  & \textbf{3.57$\pm$1.08(0\%)}  & 4.79$\pm$2.80(6\%)  & \textbf{0.68$\pm$0.07(8\%)}    & 0.67$\pm$0.03(6\%) \\
\cmidrule[\heavyrulewidth]{2-8}
\end{tabular}}
% \vspace{-4mm}
\end{table}
\begin{figure}[t]
\begin{subfigure}{.33\textwidth}
  \centering
  \includegraphics[width=1\linewidth]{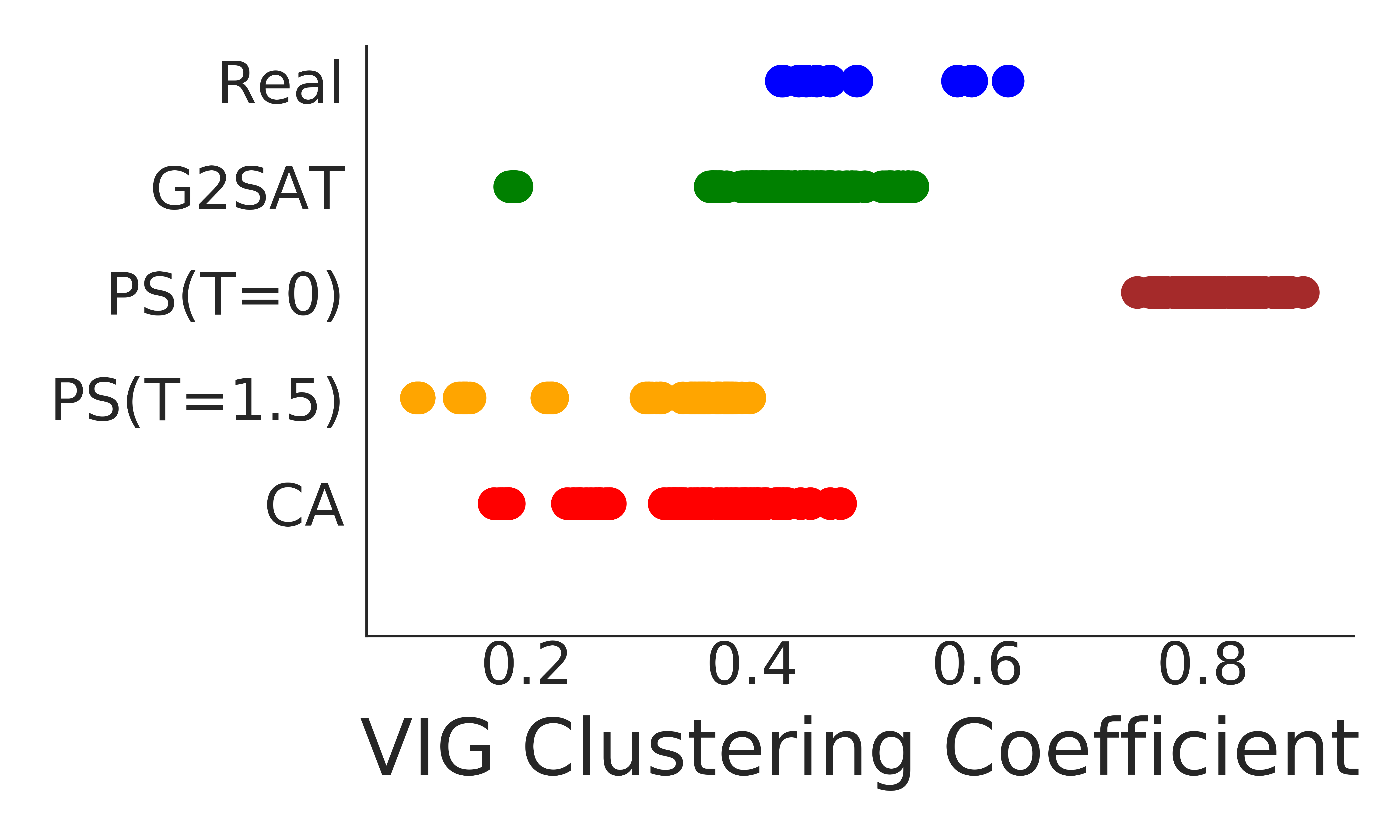}
%   \caption{1a}
  \label{fig:sfig1}
\end{subfigure}%
\begin{subfigure}{.33\textwidth}
  \centering
  \includegraphics[width=1\linewidth]{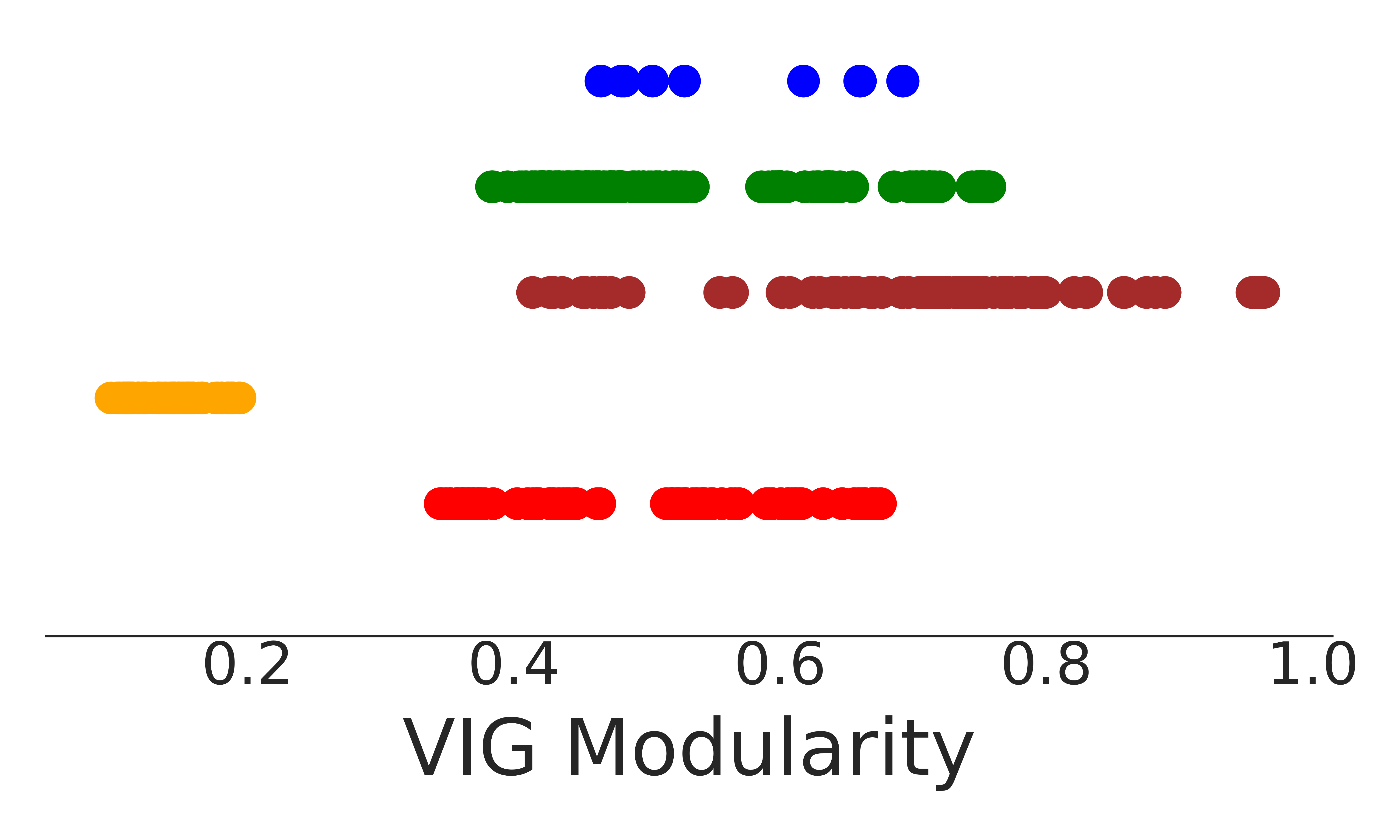}
%   \caption{1b}
  \label{fig:sfig2}
\end{subfigure}
\begin{subfigure}{.33\textwidth}
  \centering
  \includegraphics[width=1\linewidth]{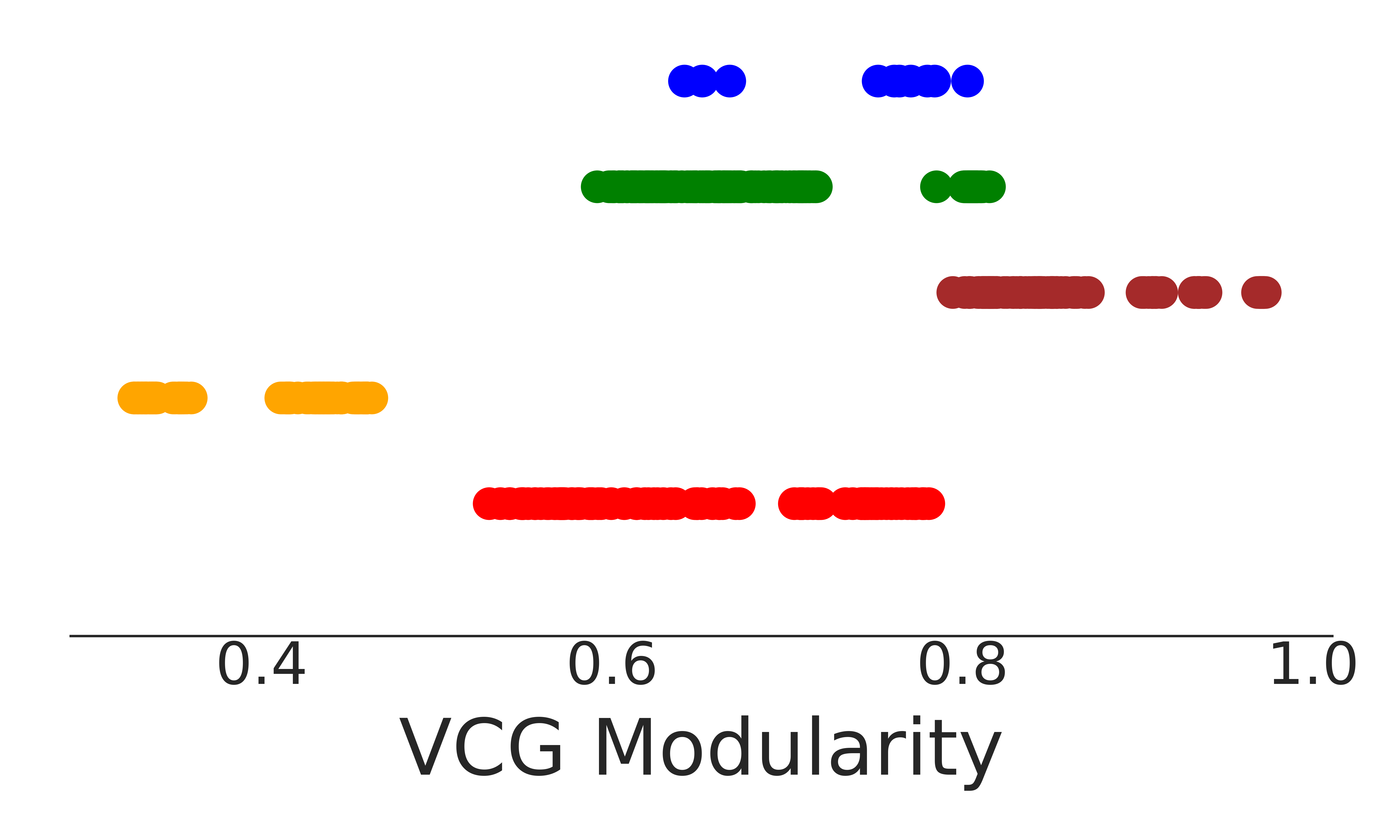}
%   \caption{1b}
  \label{fig:sfig3}
\end{subfigure}
\vspace{-4mm}
\caption{Scatter plots of distributions of selected properties of the generated formulas.}
\label{fig:stats}
% \vspace{-4mm}
\end{figure}

\xhdr{Graph statistics} 
The graph statistics of the generated SAT formulas are shown in Table \ref{table:stats}. We observe that G2SAT is the only model that is able to closely fit \emph{all} the graph properties that we measure, whereas the baseline models only fit some of the statistics and fail to perform well on the other statistics. Surprisingly, G2SAT fits the modularity even better than CA, which is tailored for fitting that statistic. We compute the relative error over the generated graph statistics with respect to the ground-truth statistics, and G2SAT can reduce the relative error by 24\% on average compared with baseline methods. To further illustrate this performance gain, we plot the distribution of the selected properties over the generated formulas in Figure \ref{fig:stats}, where each dot corresponds to a graph. We see that G2SAT nicely interpolates and extrapolates on all the statistics of the input graphs, while the baselines only do well on some of the statistics.

\xhdr{SAT solver performance}
As seen in Table \ref{table:solver_rank}, the ranking of solver performance over the formulas generated by G2SAT and CA align perfectly with their ranking over the training graphs. Both models are able to correctly
%correctly let application-focused solvers ($I_1, I_2, I_3$) to have better performance then random-focused solvers ($R_1, R_2, R_3$), 
generate formulas on which application-focused solvers ($I_1, I_2, I_3$) outperform random-focused solvers ($R_1, R_2, R_3$). %, achieving 100\% accuracy.
By contrast, PS models do poorly at this task. %More interestingly, the solver ranking induced by G2SAT aligns better with the 2018 SAT competition results when compared to using the training SAT formulas.
%\jure{We have to comment/explain why I3 is better than I1,I2 for G2SAT. We need one sentence here}
% and in the training set, were able to make a clear distinction between two genres of solvers, while formulas generated by PS failed to do so.

\xhdr{Application: Developing better SAT solvers}
The run time gain of tuning solvers on synthetic formulas compared to tuning on a small set of real-world formulas is shown in Table \ref{table:solver_tune}. While all the generators are able to improve the SAT solver's performance by suggesting different hyperparameter configurations, G2SAT is the only method that finds %a robust configuration
one that results in a large performance gain (18\% faster run time) on unobserved SAT formulas. 
% While in this experiment, formulas that we trained on came from different application domains.
Although this experiment is limited in scale, the promising results indicate that G2SAT could open up opportunities for developing better SAT solvers, even in application domains where benchmarks are scarce.

%Training & $I_1, I_2, I_3, R_1, R_2, R_3$ & 100\% \\ \cmidrule{1-3} CA & $I_1, I_2, I_3, R_1, R_2, R_3$ & \textbf{100\%} \\ PS(T=0) & $R_2, I_2, R_1, I_1, I_3, R_3$ & 33\% \\ PS(T=1.5) & $R_2, R_1, I_2, I_3, I_1, R_3$ & 33\% \\ \name~ & $I_3, I_1, I_2, R_1, R_2, R_3$ &%

\begin{table}[t]
\begin{minipage}{.49\textwidth}
\setlength\tabcolsep{2pt}
\centering		
\caption{Relative SAT Solver Performance\\ on training
as well as synthetic SAT formulas.} 
\label{table:solver_rank}
\begin{tabular}{ccc}\\
\cmidrule[\heavyrulewidth]{1-3}
Method    & Solver ranking                & Accuracy \\ \cmidrule{1-3}
Training  & $I_2, I_3, I_1, R_2, R_3, R_1$ & 100\% \\ \cmidrule{1-3}
CA        & $I_2, I_3, I_1, R_2, R_3, R_1$ & \textbf{100\%} \\ 
PS(T=0)   & $R_3, I_3, R_2, I_2, I_1, R_1$ & 33\%  \\ 
PS(T=1.5) & $R_3, R_2, I_3, I_1, I_2, R_1$ & 33\%  \\ 
\name~    & $I_1, I_2, I_3, R_2, R_3, R_1$ & \textbf{100\%}  \\ \cmidrule[\heavyrulewidth]{1-3}
\end{tabular}
\vspace{-3mm}
\end{minipage}
\hfill
\begin{minipage}{.49\textwidth}
\setlength\tabcolsep{2pt}
\centering		
\caption{Performance gain when using generated SAT formulas to tune SAT solvers.} 
\label{table:solver_tune}
\begin{tabular}{ccccc}\\
\cmidrule[\heavyrulewidth]{1-4}
Method    & Best parameters & Runtime(s) & Gain \\ \cmidrule{1-4}
Training  & (0.95, 0.9) & 2679 & N/A\\ \cmidrule{1-4}
CA        & (0.75, 0.99) & 2617 & 2.31\%\\
PS(T=0)   & (0.75, 0.999) & 2668 & 0.41\%  \\ 
PS(T=1.5) & (0.95, 0.9) & 2677 & 0.07\% \\ 
\name~    & (0.95, 0.99) & \textbf{2190} & \textbf{18.25\%} \\ \cmidrule[\heavyrulewidth]{1-4}
\end{tabular}
\vspace{-3mm}
\end{minipage}
\end{table}

\subsection{Analysis of Results}

\xhdr{Scalability of G2SAT}
While existing deep graph generative models can only generate graphs \begin{wrapfigure}{r}{0.4\linewidth}
  \centering
  \vspace{-10pt}
  \includegraphics[width=\linewidth]{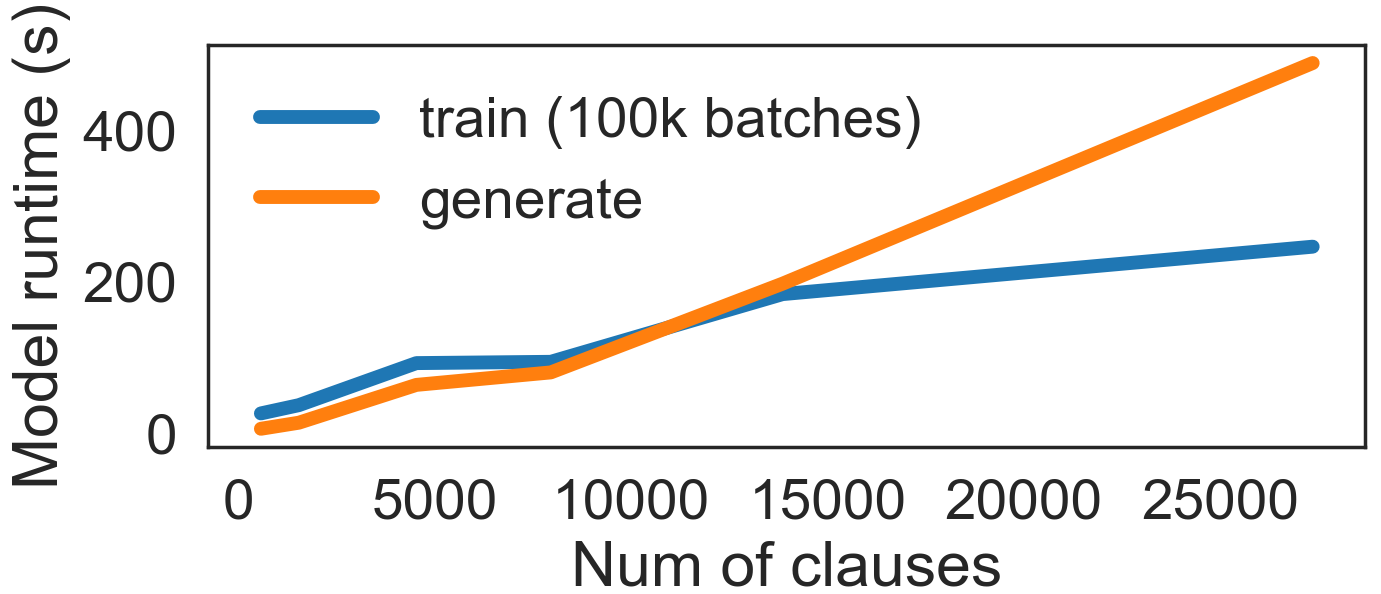}
  \vspace{-10pt}
  \caption{G2SAT Run time.}
  \label{scalability}
  \vspace{-10pt}
\end{wrapfigure}
with up to about 1,000 nodes \cite{netgan,grover2018graphite,you2018graphrnn}, the novel design of the G2SAT framework
enables the generation of graphs that are an order of magnitude larger. The largest graph we have generated has 39,578 nodes and 102,927 edges, which only took 489 seconds (data-processing time excluded) on a single GPU. Figure \ref{scalability} shows the time-scaling behavior for both training (from 100k batches of node pairs) and formula generation. We found that G2SAT scales roughly linearly for both tasks with respect to the number of clauses. 

\xhdr{Extrapolation ability of G2SAT}
To determine whether a trained model can learn to generate 
SAT instances different from those in the training set, we design an extrapolation experiment as follows. We train on 10 small formulas with 327 to 4,555 clauses, while forcing G2SAT to generate large formulas with 13,028 to 27,360 clauses.  We found that G2SAT can generate large graphs whose characteristics are similar to those of the small training graphs, which shows that G2SAT has learned non-trivial properties of real-world SAT problems, and thus can extrapolate beyond the training set. Specifically, the VCG modularity of the large formulas generated by G2SAT is 0.81 $\pm$ 0.03, while the modularity of the small formulas used to train G2SAT is 0.74 $\pm$ 0.06.

\xhdr{Ablation study} Here we demonstrate that the expressive power of GCN model significantly
\begin{wrapfigure}{r}{0.32\linewidth}
  \centering
  \vspace{-14pt}
  \includegraphics[width=\linewidth]{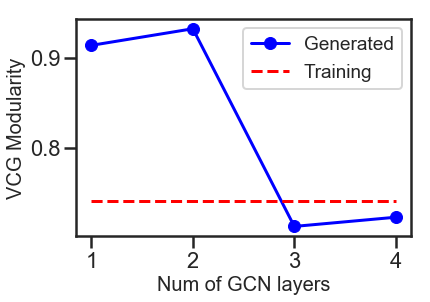}
  \vspace{-14pt}
  \caption{Ablation study.}
  \label{ablation}
  \vspace{-30pt}
\end{wrapfigure}
affects the generated formulas. Figure \ref{ablation}
shows the effect of the number of layers in the GCN neural network model on the modularity of the generated formulas. As the number of layers increases, the average modularity of the generated formulas becomes closer to that of the training formulas, which indicates that machine learning contributes significantly to the efficacy of G2SAT. The other graph properties that we measured generally follow the same pattern as well.
% \begin{wrapfigure}{R}{0.32\linewidth}
%   \centering
%   \vspace{-19pt}
%   \includegraphics[width=\linewidth]{LaTeX/figs/ablation.png}
%   \vspace{-10pt}
%   \caption{Ablation Study}
%   \label{ablation}
%   \vspace{-10pt}
% \end{wrapfigure}

\section{Conclusions}
\label{sec:concl}

In this paper, we introduced G2SAT, the first deep generative model for SAT formulas. In contrast to existing SAT generators, G2SAT does not rely on hand-crafted algorithms and is able to generate diverse SAT formulas similar to input formulas, as measured by many graph statistics and SAT solver performance. While future work is called for to generate larger and harder formulas, we believe our framework shows great potential for understanding and improving SAT solvers.

% At the highest level, SAT-GEN reads a real-world SAT-formula and generates random formulas that mimic it. Internally, SAT-GEN first transforms the formula to its LIG, then uses an implicit graph modelling technique, NetGAN, to generate realistic biased random walks on the LIG. Next, using the generated random walks, the model constructs new graphs that are interpreted as LIGs. Finally, running a greedy hill-climbing algorithm for minimum clique edge cover, followed by an cover-expansion process, SAT-GEN extracts SAT-formulas from the generated graphs. 

% We have shown that \satgen~captures a wider range of properties of real-world SAT formulas than any of the previous generators. In particular, properties that have been shown to be important, such as modularity and scale-free structures are captured with greater fidelity.\\

% There are two avenues for further research. First, our way of extracting SAT formulas from an LIG sometimes fails to capture the power-law distribution of the clause degree. Second, the generated formulas are often much easier than the original real-world formulas on which they are modeled. Nevertheless, we believe that our preliminary exploration of a graph-based implicit SAT-formula generator shows great promise.

%We have discussed potential ways to tackle these two problems, which will be left as future work.

\section*{Acknowledgements}
Jure Leskovec is a Chan Zuckerberg Biohub investigator.
We gratefully acknowledge the support of 
DARPA under No. FA865018C7880 (ASED) and MSC;
NIH under No. U54EB020405 (Mobilize); 
ARO under No. 38796-Z8424103 (MURI); 
IARPA under No. 2017-17071900005 (HFC);
NSF under No. OAC-1835598 (CINES) and HDR;
Stanford Data Science Initiative, 
Chan Zuckerberg Biohub,  Enlight Foundation,
JD.com, Amazon, Boeing, Docomo, Huawei, Hitachi, Observe, Siemens, UST Global.
The U.S. Government is authorized to reproduce and distribute reprints for Governmental purposes notwithstanding any copyright notation thereon. Any opinions, findings, and conclusions or recommendations expressed in this material are those of the authors and do not necessarily reflect the views, policies, or endorsements, either expressed or implied, of DARPA, NIH, ONR, or the U.S. Government.

\section{Appendix}

\subsection{Computing facilities}
The G2SAT model is trained on a single NVIDIA RTX-2080Ti GPU.
We evaluate the performance of SAT solvers on a cluster equipped with Intel Xeon E5-2637 v4 CPUs running Ubuntu 16.04 and we dedicated 2 cores, 8000 MB RAM for each job. 
For each formula, we gave each solver a 10 minutes timeout.

\subsection{Evaluation}
We used an implementation of the Louvain Algorithm to measure the modularities \cite{louvain}.

We used an implementation of the maximum likelihood method 
% \footnote{\url{http://www.iiia.csic.es/~levy/software/scalefree.cpp}}
for computing an estimate of $\alpha_v$ and $\alpha_c$ \cite{max-likely, scale-free}.

\subsection{Removing Trivial Components of SAT formulas Generated by PS}
% \subsection{Post-processing}

We found that almost all formulas generated by the PS model were trivially unsatisfiable when it is tasked with generated formulas with similar metrics as formulas in the traning set. In order to better demonstrate the SAT solver behaviors on the PS model, we increased the difficulty of its generated formulas by performing a lightweight post-processing step that iteratively removes clauses leading to short unsatisfiable proof. Concretely, for each generated formula, we used a SAT solver (Picosat \cite{picosat}) to try to solve it within a given conflict\footnote{A SAT solver finds a conflict when it finds a partial assignment of values to variables that cannot satisfy the formula.} budget. If the solver were able to prove that the formula is unsatisfiable within that budget, we removed a clause from the unsatisfiable core (i.e., a subset of all clauses that is unsatisfiable) that Picosat returned and repeated the same process on the pruned formula, until it became satisfiable or no longer solvable within the conflict budget. We found that this efficient approach, only removes a low proportion of clauses and has small impact on the graph theoretic properties, but significantly increases the difficulty of the formulas generated by PS.

\subsection{More details on baseline methods}
\xhdr{Community Attachment (CA)}
The CA model generates formulas to fit a desired VIG modularity \cite{mod_gen}. 
The model takes in five inputs $n, m, k, c, Q$, where $n$ is the number of variables, $m$ the number of clauses, $k$ the length of each clause, $c$ the size of a partition of the VIG, and $Q$ is the desired VIG modularity. 
The output of the algorithm is a SAT formula with $n$ variables and $m$ clauses, each of length $k$, such that the optimal modularity for any $c$-partition of the VIG of the formula is approximately $Q$.

% The model takes in five inputs $n, m, k, c, Q$, where $n$ is the number of variables, $m$ the number of clauses, $k$ the length of each clause, $c$ the size of a partition of the VIG, and $Q$ is the desired VIG modularity. The output of the algorithm is a SAT formula with $n$ variables and $m$ clauses, each of length $k$, such that the optimal modularity for any $c$-partition of the VIG of the formula is approximately $Q$.

\xhdr{Popularity-Similarity (PS)}
% The PS model generates formulas to fit desired $\alpha_v$ and $\alpha_c$ \cite{loc_gen}. 
% $T$ is a hyper-parameter that decides the trade-off between modularity and $\alpha_v, \alpha_c$ of generated formulas. We use two versions of PS, with $T=0$ and $T=1.5$.
The PS model generates formulas to fit desired $\alpha_v$ and $\alpha_c$ \cite{loc_gen}. 
In addition, the formulas generated by PS are guaranteed to have high modularity. 
The model takes in seven inputs $n, m, k, K, \alpha_v, \alpha_c, T$, where $n$, $m$ are the same with CA, $k$ the minimum clause length, $K$ the average clause length, and $T$ a hyper-parameter that decides the trade-off between modularity and $\alpha_v, \alpha_c$. We use two versions of PS, with $T=0$ and $T=1.5$.

\bibliography{bibli}
\bibliographystyle{abbrv}

\end{document}